\documentclass[conference]{IEEEtran}
\IEEEoverridecommandlockouts

\usepackage{cite}
\usepackage{amsmath,amssymb,amsfonts}
\usepackage{algorithmic}
\usepackage{graphicx}
\usepackage{textcomp}
\usepackage{xcolor}
\def\BibTeX{{\rm B\kern-.05em{\sc i\kern-.025em b}\kern-.08em
    T\kern-.1667em\lower.7ex\hbox{E}\kern-.125emX}}
\begin{document}

\title{Generating Realistic COVID19 X-rays with a \\ Mean Teacher + Transfer Learning GAN\\
\thanks{This research was supported by NSF award titled {\em RAPID: Deep Learning Models for Early Screening of COVID-19 using CT Images}, award \# 2027628}
}

\author{Sumeet Menon$^1$, Joshua Galita$^1$, David Chapman$^1$, Aryya Gangopadhyay$^1$, Jayalakshmi Mangalagiri$^1$, Phuong Nguyen$^1$  \\ \\
Yaacov Yesha$^1$, Yelena Yesha$^1$, Babak Saboury$^1$$^,$$^2$, Michael Morris$^1$$^,$$^2$$^,$$^3$ \\
\\ 
\normalsize{\quad \quad \quad  $^1$University of Maryland, Baltimore County \quad \quad \quad \quad \quad $^2$National Institutes of Health Clinical Center }\\
\normalsize{1000 Hilltop Circle, Baltimore, MD, 21250 \quad \quad \quad \quad \quad \quad 9000 Rockville Pike,  MD 21201}\\
\normalsize{\quad \quad sumeet1@umbc.edu} \quad \quad \quad \quad \quad \quad \quad \quad \quad \quad \quad \quad \quad \quad \quad \quad \quad \quad \quad \quad \quad \quad \quad \quad \\ \normalsize{\quad \quad $^3$Networking Health}\\ \normalsize{\quad \quad 331 Oak Manor Drive, Suite 201, Glen Burnie MD, 21061}}

\maketitle

\begin{abstract}
COVID-19 is a novel infectious disease responsible for over 800K deaths worldwide as of August 2020.  The need for rapid testing is a high priority and alternative testing strategies including X-ray image classification are a promising area of research.   However, at present, public datasets for COVID19 x-ray images have low data volumes, making it challenging to develop accurate image classifiers.  Several recent papers have made use of Generative Adversarial Networks (GANs) in order to increase the training data volumes.  But realistic synthetic COVID19 X-rays remain challenging to generate.  We present a novel Mean Teacher + Transfer GAN (MTT-GAN) that generates COVID19 chest X-ray images of high quality.  In order to create a more accurate GAN, we employ transfer learning from the Kaggle Pneumonia X-Ray dataset, a highly relevant data source orders of magnitude larger than public COVID19 datasets.  Furthermore, we employ the Mean Teacher algorithm as a constraint to improve stability of training.  Our qualitative analysis shows that the MTT-GAN generates X-ray images that are greatly superior to a baseline GAN and visually comparable to real X-rays.  Although board-certified radiologists can distinguish MTT-GAN fakes from real COVID19 X-rays. Quantitative analysis shows that MTT-GAN greatly improves the accuracy of both a binary COVID19 classifier as well as a multi-class Pneumonia classifier as compared to a baseline GAN.  Our classification accuracy is favourable as compared to recently reported results in the literature for similar binary and multi-class COVID19 screening tasks.

\end{abstract}

\begin{IEEEkeywords}
Coronavirus, deep transfer learning, mean teacher.
\end{IEEEkeywords}

\section{Introduction}

This novel virus was reported to have originated from Wuhan, Hubei province, China in 2019. This disease is transmitted by inhalation or contact with infected droplets and the incubation period ranges from 2 to 14 days \cite{b1}. In the study of a patient who was a worker at the market and  was admitted to the Central Hospital of Wuhan on 26 December 2019, the patient was reported to be experiencing a severe respiratory syndrome that included fever, dizziness and a cough which proved to be one of the major symptoms of the virus. The complete 16 biological analysis stated that the virus showed similarities to a group of SARS-like coronaviruses which was previously found in bats in China \cite{b2},\cite{b3}. Across 150 states, over 750,000 individuals were reported to be infected by SARS-CoV-2 with a death rate of 4\% \cite{b4},\cite{b5}. 

Rapid testing is a major need across the world. The primary testing modality is molecular testing \cite{kubina2020molecular} of which nucleic acid testing for discriminating genes is the dominant approach.  However, a challenge is that nucleic acid testing requires culturing a sample which can take several days.  An alternative is \textit{rapid} serological testing \cite{bastos2020diagnostic} which detects COVID19 antigens.  However, rapid serological testing is not intended, and may be less effective, for detecting the currently infected individuals.  It is nevertheless widely used in hospitals even for this purpose due to its rapid turnaround time despite the high potential for false negatives \cite{cassaniti2020performance}. An alternative modality that is often employed in hospitals is to make use of X-ray or CT-scan imaging to detect the presence of traces of pneumonia \cite{zhou2020ct, shi2020radiological}.  For human radiologists, CT scans are a modality that offer a high discriminating power of the disease at early stages \cite{zhou2020ct}.  Chest X-rays are also used, but X-rays are often difficult to interpret as several indicators of COVID-19 infection including Ground Glass Opacity (GGO) are more difficult to discern by eye in X-rays versus CT scans by human Radiologists \cite{shi2020radiological}. 

Many recent papers have attempted to develop a deep learning classifier for COVID19 using X-ray imagery \cite{hemdan2020covidx, sethy2020detection, wang2020covid, ozturk2020automated}.  Image classification has the potential to provide immediate testing results, by identifying distinguishing imaging biomarkers.
Image classification algorithms for COVID19 have relied heavily on public datasets;  in particular the covid-chestxray-dataset in conjunction with the Kaggle Pneumonia competition dataset \cite{cohen2020a, cohen2020b, kaggle}.  Due to the availability of these datasets, it is not uncommon for investigators to construct a multi-class classifier to distinguish Normal X-rays, Bacterial Pneumonia, Viral Pneumonia and COVID19.  However, public availability of COVID19 X-ray datasets have limited data volumes numbering in low hundreds of images.  As such, several recent papers have investigated ways of increasing data volumes by making use of Generative Adversarial Networks (GANs) for deep augmentation \cite{b5, b20}.

To the best of our knowledge, no GAN algorithms for COVID19 chest X-rays, including ours, have achieved clinical quality for use by human Radiologists.  Yet, improvements to image quality translate to improved classification accuracy for automated screening algorithms.  A difficulty with GAN generated COVID19 X-rays is the presence of fuzzy boundaries over major anatomical features such as heart, liver, and ribcage \cite{b5, b20}.  Fuzzy image quality for generated X-rays is attributable to insufficient data volumes of COVID19 images.  Nevertheless, these deep generated X-rays still yield discriminating features and improve classification accuracy \cite{b5, b20}.  One of the most discriminating features of COVID19 is the presence of GGO that can be observed as large regions of the lung that are lighter and somewhat more opaque relative to normal healthy lungs \cite{zhou2020ct,shi2020radiological}.  This GGO may be visible to deep learning even in images that cannot reliably generate crisp boundaries around organs and rib-cage.

Ideally, however, the generated COVID19 X-rays should achieve the highest quality possible such as to approximate the real COVID19 images.  For example, if the generated COVID19 images are overly fuzzy, then a classifier might mistakenly learn that the fuzzy images are indicative of COVID19, but the sharply defined images are non-COVID19.

The proposed MTT-GAN architecture greatly improves image quality through transfer learning from the Kaggle Pneumonia dataset.
In order to further improve the accuracy for screening
we propose to combine this transfer learning with an exponential moving average approach based on the mean teacher algorithm.
Although transfer learning from ImageNet is widely adopted and employed for COVID19 classification, ImageNet is not an X-ray dataset.  MTT-GAN is unique in that it employs transfer learning to both the generator and discriminator from the Kaggle Pneumonia X-ray dataset not ImageNet.
The Kaggle Pneumonia X-ray dataset is close to the target domain and is orders of magnitude larger than the covid-chestxray-dataset, thereby making an ideal data source for transfer learning to COVID19.

MTT-GAN is also unique because it employs the exponential moving average component of the mean teacher algorithm \cite{b7}. Mean Teacher combines two models: a student and teacher in which the teacher performs exponential moving average of student weights in order to estimate an improved gradient direction.
The mean teacher algorithm makes the gradient descent converge more consistently and to a better global optimum than Adam optimization alone for both fully supervised and semi-supervised models \cite{b7}.

\begin{figure}[ht]
    \includegraphics[width=90mm]{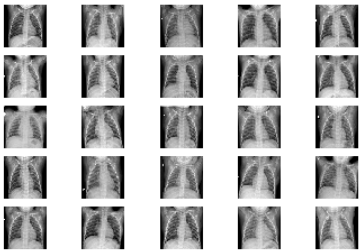}
    \caption{Kaggle Pneumonia/Normal Chest X-rays}
    \label{fig:Kaggle}
\end{figure}

\begin{figure}[ht]
    \includegraphics[width=90mm]{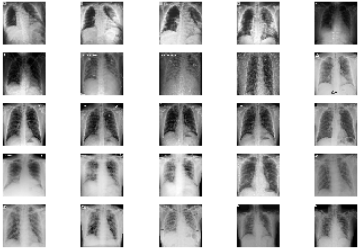}
    \caption{COVID-19 Chest X-rays}
    \label{fig:Covid}
\end{figure}

\section{Related Work}

Perhaps the most similar recent work to ours is that of Loey et al. (2020) \cite{b5} which  employs conditional GAN augmentation to improve the accuracy of multiclass classification to identify COVID19 versus normal, bacterial pneumonia and viral pneumonia as well as binary classification between COVID19 and normal X-rays \cite{b5}.  The testing accuracy on 4 classes (covid, normal, bacterial pneumonia and viral pneumonia) was 66.7\%, 80.56\% and 69.46\% using AlexNet, GoogleNet and ResNet18 respectively \cite{b5}.  We have trained and tested our model on a comparably large amount of data with 4 classes and we achieve a test accuracy of 83.45\% and 84.91\% on VGG-19 and AlexNet respectively.

A notable distinction between our approach and Loey et al. (2020), is that we propose the use of transfer learning from Kaggle pneumonia to train the GAN generator and discriminator models, rather than only the final classification step \cite{b5}.  As such, MTT-GAN is capable of generating higher quality images with more anatomical detail.

Narin et al. (2020) also generate synthetic COVID19 X-rays using GANs \cite{b20}. The authors compare 3 different generator architectures viz., ResNet50, InceptionV3 and InceptionResNetV2, of which ResNet50 provides the best accuracy. Binary classification is performed between the COVID19 and the normal X-rays by training and testing (10 COVID19 + 10 normals X-rays) on a relatively small data-set.  \cite{b20}.
Similarly to Narin et al. (2020) we make use of a ResNet like architecture for the generator.  We extend this approach by incorporating Mean Teacher and transfer learning from the Kaggle pneumonia dataset.  Our classification results are comparable but evaluated with a larger testing dataset \cite{b20}.

The use of GANs to improve pulmonary disease classification with X-rays was first performed in 2018 simultaneously by Salehinejad et al. (2018) and Madani et al. (2018) \cite{salehinejad2018generalization, madani2018chest}.
Salehinejad et al. (2018) generate chest X-rays to improve the performance of a classifier model for five categories of lung diseases \cite{salehinejad2018generalization}.
Qualitatively, a board certified radiologist was able to identify the features pertaining to the five categories in the generated images.  Quantitatively, the use of the GAN to augment their dataset improved the performance of the classifier.
In the same year Madani et al. (2018) trained two deep convolutional GANs to generate normal X-rays and X-rays with cardiovascular abnormalities \cite{madani2018chest}. The authors compared the accuracy of a classifier using unaugmented training data, data augmented using traditional methods, such as shifts and cropping, and data augmented using a GAN and traditional methods.

In addition to works that have incorporated the use of GANs a variety of papers have worked on classifiers between COVID19, bacterial pneumonia, viral pneumonia, and normal healthy lungs \cite{sethy2020detection,wang2020covid}.
Sethy, et al. (2020) have created a dataset of 381 X-rays amongst 3 classes: COVID19, pneumonia and normal \cite{sethy2020detection}.
The authors compare several models and achieve the highest accuracy of 98.6\% with a ResNet50 plus SVM architecture as compared to the 93.4\% by the traditional approach \cite{sethy2020detection}. 
Wang, L. et al. (2020) have investigated a similar classification problem with a novel COVID-NET architecture featuring a lightweight residual projection-expansion projection-extension (PEPX) design pattern \cite{wang2020covid}.

\section{Data Preparation}

The datasets used for our study are the Kaggle pneumonia chest X-rays dataset \cite{kaggle} and the COVID19 open source chest X-rays \cite{cohen2020a, cohen2020b}. The Kaggle pneumonia dataset consists of 5,856 X-ray images (JPEG) with 3 categories (normal, bacterial pneumonia, viral pneumonia). This dataset (anterior and posterior) was taken from a collection of pediatric patients. Some sample images from the Kaggle dataset are shown in Fig. \ref{fig:Kaggle}. The COVID19 dataset, after removing images from patients without COVID19 and those from a lateral view, consists of 227 X-ray images of patients with the coronavirus. Some sample images from the COVID19 dataset are shown in Fig. \ref{fig:Covid}. For our study, we downsampled all of the images to 128x128 resolution.

Both the Kaggle pneumonia and COVID19 datasets are augmented using soft augmentation through cropping. For each original image, we generate a certain number of augmented images, which we call the augmentation factor. For each augmented image, the image is cropped on all four sides by a percentage randomly chosen between zero and five percent. Thus, at a minimum, the middle 90\% of the original image, measured in both the horizontal and vertical directions, remains. For the Kaggle pneumonia dataset, we choose an augmentation factor of 5; for the COVID19 dataset, we choose an augmentation factor of 50.

We were careful not to employ horizontal flipping.  Imposing a horizontal flip would cause the cardiac silhouette to appear on the opposite side of the body, which would be clinically incorrect for the majority of patients.

\begin{figure*}
\centerline{\includegraphics[width=150mm,height=70mm,scale=10.5]{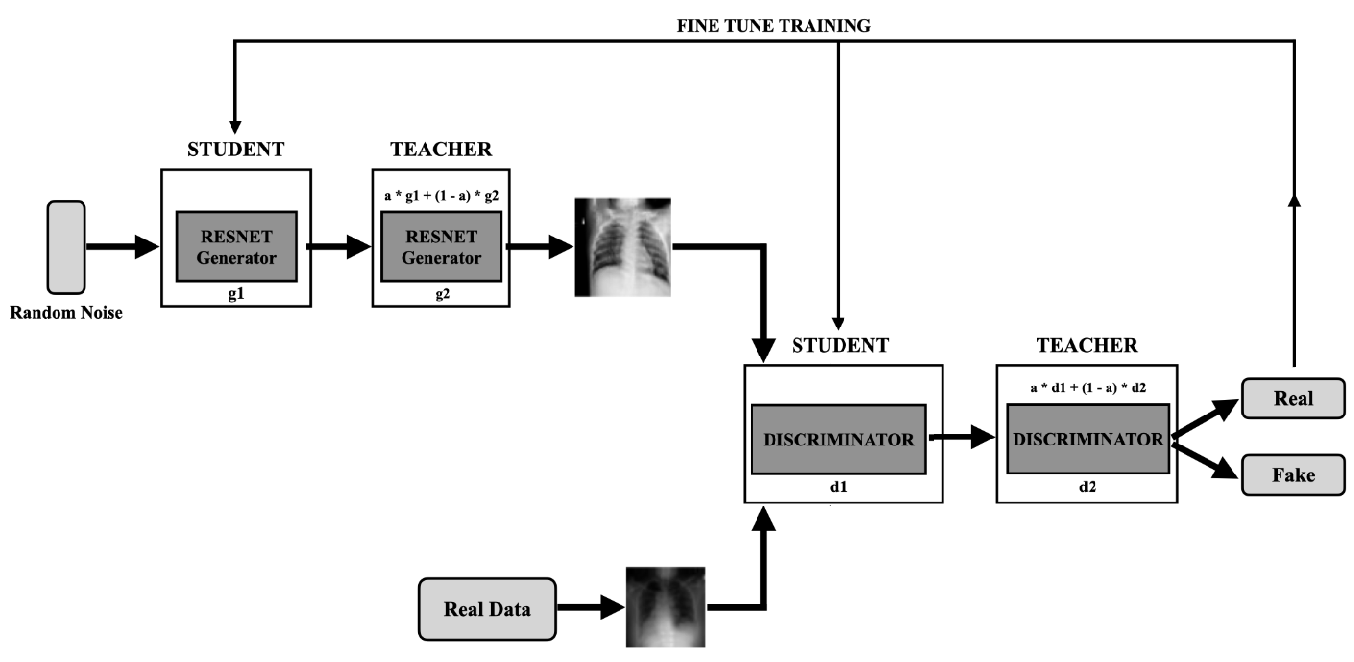}}
\caption{MTT-GAN fine-tuning algorithm showing student and teacher for generator and discriminator models.}
\label{fig:mean_teacher}
\end{figure*}

\section{Methodology}
\subsection{Transfer Learning}
Transfer learning is a method for addressing the problem of insufficient training data by using additional data from another larger source
\cite{b8, b11, b13, b14}.  Due to the low volume of COVID19 images available we wish to employ transfer learning from the Kaggle pneumonia dataset.
Intuitively, we assume that the ideal weights for the model for generating COVID19 X-rays are closer to the weights of the model after training on pneumonia and normal X-rays (the Kaggle dataset) than they are to the weights at initialization. For the most part, COVID19 X-rays are more similar to Kaggle Pneumonia X-rays than they are to white noise. Thus, we expect that the difference between the ideal COVID19 weights of the model and the weights after transfer learning to be substantially smaller than the difference between the COVID19 weights and the randomly initialized weights.

The MTT-GAN is first trained to convergence on the Kaggle dataset, and subsequently fine tuned on the COVID19 dataset.  To improve convergence, the fine tuning makes use of an exponential moving average learning algorithm based on Mean Teacher \cite{b7}.  Care was taken to ensure that the training and testing splits were completely separated, both for pre-training and fine tuning.  30\% of the real covid X-rays (i.e. 68 X-rays) were removed from the COVID19 dataset and 68 images of each class (normal, bacterial pneumonia, and viral pneumonia) from the Kaggle dataset.
For each training epoch, the training was split into mini-batches of size 32. For each mini-batch, the discriminator model was trained using 16 real images (labeled 1) and 16 fake images (labeled 0), with the fake images generated by passing Gaussian noise on the interval [0, 1] through the generator model. Then, the combined model, consisting of both the discriminator and the generator, was trained using 32 noise vectors.
A crossentropy loss was used with the Adam optimizer using a learning rate of $10^{-5}$ and a $\beta_1$ of 0.5. The training was run for 100 epochs using the Kaggle dataset followed by 100 epochs with the exponential moving average algorithm using the COVID19 dataset.

\subsection{Exponential Moving Average Training}

MTT-GAN employs a supervised version of the Mean Teacher algorithm featuring exponential moving average of model weights in order to improve training convergence of the generator and discriminator as seen in figure \ref{fig:mean_teacher}.  Mean Teacher is equally applicable to both semi-supervised as well as supervised learning, and is widely regarded as a state-of-the-art semi-supervised learning approach \cite{b7}.
Mean teacher employs two models in parallel, a student and a teacher model. After each gradient step, The student model performs gradient descent, but the teacher model is updated to use an exponential moving average of the student weights, Due to the use of this exponential moving average, the teacher model is expected to converge faster than the student model.  As such, the teacher model is ultimately used for classification  \cite{b7}.  

The training loss of MTT-GAN differs from the original Mean Teacher, because MTT-GAN does not include a consistency loss between pseudo labels of the teacher models and the predicted labels of the student models.  The original Mean Teacher incorporates this consistency loss, in which the teacher model predicts the labels of unlabeled samples, and the student model performs an additional gradient descent toward the teacher's predictions.  However, this step is not necessary for MTT-GAN, because the minimax loss function is completely supervised \cite{b7}.  

The student discriminator loss is defined as follows,

\begin{equation}
    L_D = -\mathbf{E}_{x_r\in{S}} \log(D(x_r)) 
    - \mathbf{E}_{z\in{Z}} \log(1 - D(G(z)))
\end{equation}

where $S$ is the set of real images and $Z$ is the latent space from which noise, $z$, taken. On the other hand, the student generator, $G$, tries to optimize its weights to maximize that same equation, or equivalently, minimize the equation.

\begin{equation}
    L_G = -\mathbf{E}_{z\in{Z}} \log(D(G(z))) 
    - \mathbf{E}_{x_r\in{S}} \log(1 - D(x_r))
\end{equation}

After each gradient step of the student generator and student discriminator models as seen in figure \ref{fig:mean_teacher}, the respective teacher's weights are updated using an exponential moving average of the student weights as follows,

\begin{equation}
    \theta^{'}_t  = \alpha \; \theta^{'}_{t-1}  +  (1-\alpha) \; \theta_t
\end{equation}

Where $\theta^{'}_t$ represents the weights of the respective teacher model at time-step $t$, whereas $\theta_t$ represents the weights of the respective student model at time-step $t$.

\subsection{GAN Architecture}

\begin{figure*}
    \centerline{\includegraphics[width=\linewidth]{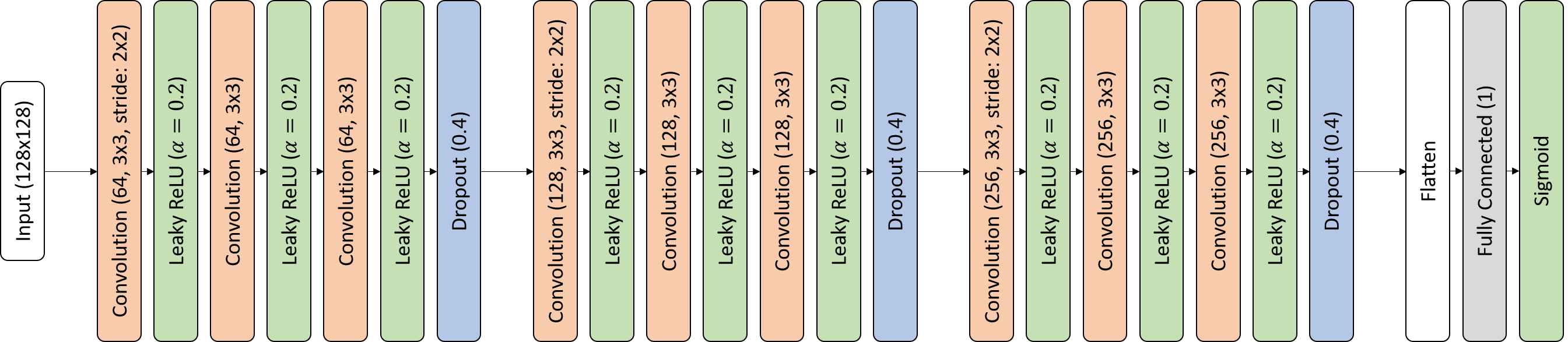}}
    \caption{Discriminator Architecture}
    \label{fig:D_arch}
\end{figure*}

The MTT-GAN architecture consists of two separate models, a generator network and a discriminator network. Our discriminator network, shown in Fig. \ref{fig:D_arch}, consists of a total of nine convolutional layers, each with a kernel size of 3x3, and each followed by a LeakyRelu activation function with an alpha value of 0.2. The first three layers have 64 filters, the next three have 128 filters, and the last three have 256. The first, fourth, and seventh layers have a stride of 2x2, the other six layers have a 1x1 stride. Additionally, a dropout layer with a dropout rate of 0.4 is included after every third layer. The last convolutional layer is fully connected to a layer with one node, and the sigmoid activation function is used.

\begin{figure*}
    \centerline{\includegraphics[width=\linewidth]{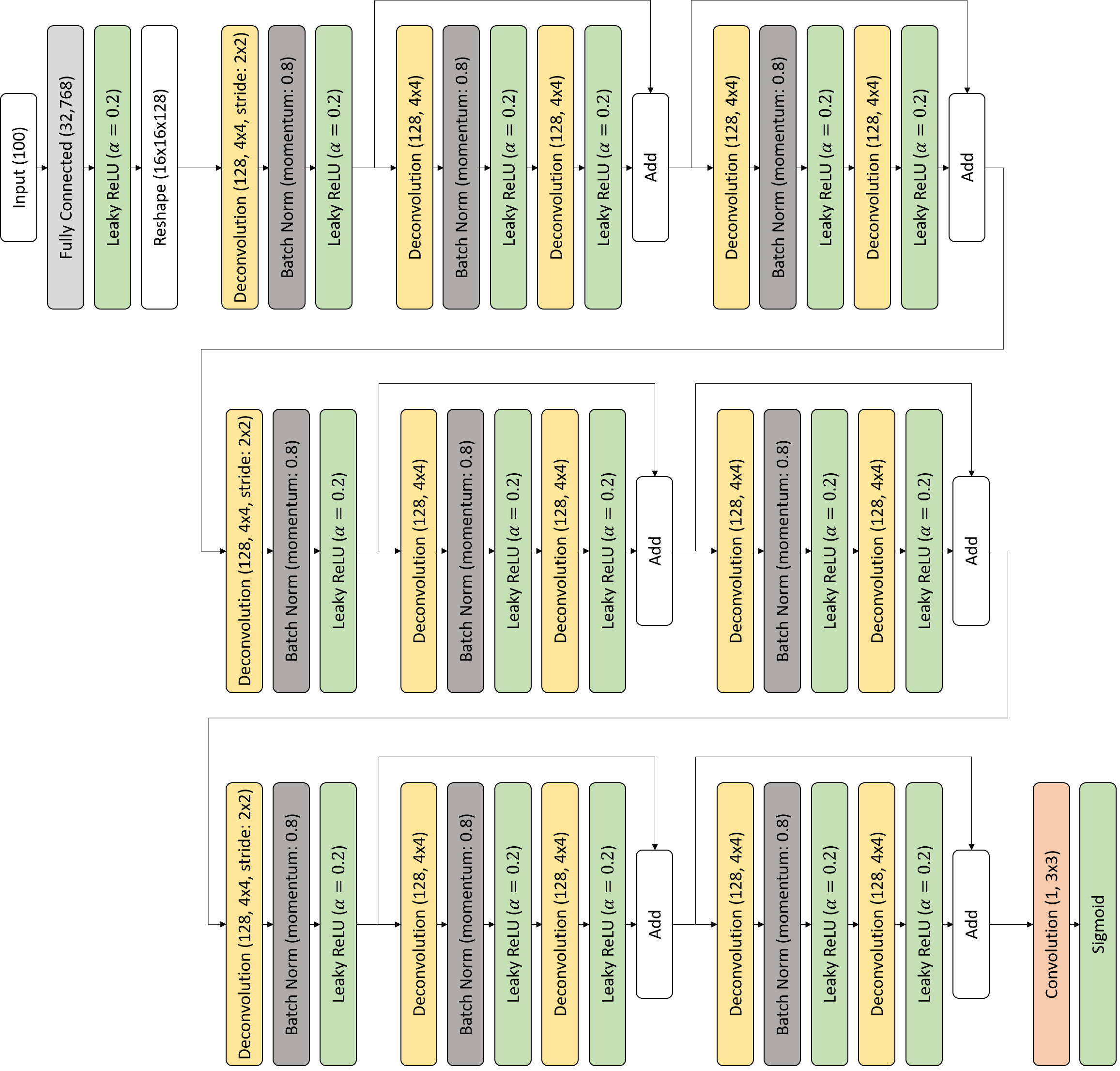}}
    \caption{Generator Architecture}
    \label{fig:G_arch}
\end{figure*}

The generator network, shown in Fig. \ref{fig:G_arch}, takes a vector in 100-dimensional latent space. This is followed by a fully connected layer that is shaped into 128 feature maps of size 16x16. The generator contains three blocks, each of which increases the size of the feature maps. Each of these blocks starts with a deconvolutional layer with 128 filters, a 4x4 kernel, and a 2x2 stride, doubling both dimensions of the feature map. Each of these upscaling deconvolutional layers is followed by two residual blocks. Each residual block consists of a deconvolutional layer (128 filters, 4x4 kernel), a batch normalization layer with a momentum of 0.8, a leakyRelu activation function, another deconvolutional layer (same specifications), and another leakyRelu activation. After all three different-sized blocks, a convolutional layer with a 3x3 kernel is used with a sigmoid activation. This generates a 128x128 image with the pixel intensities normalized to the interval [0, 1]. The loos function used was binary cross-entropy and we used the adam optimizer with a learning rate of 0.00001 and the size of each mini-batch was 32.

Residual networks, or ResNets, allow for a model to be deeper without the gradients vanishing or exploding by creating "skip connections," where a portion of the model learns the difference between the output and the input rather than learning the output from scratch \cite{b15}. ResNets have been used extensively, including with similar datasets as in Wang et al. (2020), where a residual architecture is used in a classification model for COVID19 X-ray images \cite{b19}. While first used in CNNs, Gulrajani et al. (2017) demonstrated that residual architectures can be used to improve the performance of a GAN \cite{b18}. Since then, ResNets have become commonly used in the generator portion of a GAN, such as in Ledig et al. (2017), where a ResNet is used in the generator of a super-resolution GAN. As such, we employ a residual network in our generator architecture\cite{b16}.

The discriminator network, shown in Fig. \ref{fig:D_arch}, consists of a total of nine convolutional layers, each with a kernel size of 3x3, and each followed by a LeakyRelu activation function with an alpha value of 0.2. The first three layers have 64 filters, the next three have 128 filters, and the last three have 256. The first, fourth, and seventh layers have a stride of 2x2, the other six layers have a 1x1 stride. Additionally, a dropout layer with a dropout rate of 0.4 is included after every third layer. The last convolutional layer is fully connected to a layer with one node, and the sigmoid activation function is used.

\section{Experimental Design}

The MTT-GAN architecture was evaluated by observing the effect on a classifier model when the training dataset is augmented using images generated by the GAN. Two classifier models were used for our evaluation, VGG-19 \cite{VGG} and AlexNet \cite{AlexNet}. The architectures of the models were modified only to change the number of outputs to two and four for the binary and multi-class classification problems respectively. For all experiments, the classifiers were trained from scratch for 50 epochs. The adam optimizer was used with a learning rate of $10^{-5}$. For classifier training, 30\% of the training data was reserved for validation.

Further, the MTT-GAN was compared against a Transfer-GAN and a baseline GAN. The baseline GAN has the same generator and discriminator architectures, but does not use the mean-teacher exponential moving average or the transfer learning.  The transfer-GAN is similar but does not incorporate the exponential moving average training.
All three GANs were trained using a crossentropy loss and the adam optimizer with a learning rate of $10^{-5}$. Both GANs were trained for 100 epochs on the specified COVID19 dataset, and the MTT-GAN is trained on the transfer dataset prior to the COVID19 training.

For all experiments, 68 images of each class are withheld for testing. For the binary classifier, the classes are COVID19 and normal, and for the multi-class classifier, the classes are COVID19, normal, bacterial pneumonia, and viral pneumonia. The images withheld for testing are not used in any way for training the GAN models or the classifier models.

The training dataset consists of 1400 images of each class prior to the validation split (30\% of the training data is used for validation). Thus, a total of 2800 images are used for training the binary classifier and 5600 images are used for the multi-class classifier. For the normal, bacterial pneumonia, and viral pneumonia classes, the images used for training are all real images taken from the Kaggle dataset. For the COVID19 class, the composition of the images varies between experiments. Four experiments are run on each of the classifier. In the first two experiments, the COVID19 images are images generated by the baseline GAN model. In the first, data augmentation is not used for training the GAN, while in the second, data augmentation is used, using an augmentation factor of 50 (the same as in training the MTT-GAN on COVID19 images). In the third experiment, the COVID19 images are images generated by the Transfer-GAN model. In the fourth experiment, the 1400 images consist of the 159 real COVID19 images (the remaining images after 68 were reserved for testing) and 1241 images generated using the Transfer-GAN.  The fifth and sixth experiments are similar but employ the full MTT-GAN with exponential moving average training and transfer learning. For each experiment, we report the testing accuracy, and for the experiments with only GAN-generated images, we also report a confusion matrix.  Furthermore, confidence intervals for the reported accuracy numbers are included in Tables \ref{tab:binary} and \ref{tab:multiclass} using the Clopper-Pearson method.

\section{Quantitative Results}

\begin{table}[htbp]
\label{MultiClass Classifier Experiments}
\caption{Binary Classifier Experiments}
\begin{center}
\begin{tabular}{|p{2.9cm}|p{2.1cm}|p{2.1cm}|}
\hline

\textit{\textit{Training Dataset (2800 X-rays = 1400 covid X-rays+1400 normal X-rays)}} & \textit{\textit{VGG-19 Test Accuracy with confidence intervals (136 X-rays = 68 covid X-rays+68 normal X-rays) }} & \textit{\textit{AlexNet Test Accuracy with confidence intervals (136 X-rays = 68 covid X-rays+68 normal X-rays) }} \\
\hline
\textit{\textit{Baseline Method (without augmentation) with only Generated Images}} &\textbf{74.26\%} (0.66068,0.81374)  &\textbf{77.20\%} (0.69233,0.83957)  \\
\hline
\textit{\textit{Baseline Method (with augmentation) with only Generated Images}} &\textbf{91.91\%} (0.85989,0.95893) &\textbf{82.35\%} (0.74890,0.88354)  \\
\hline
\textit{\textit{Transfer GAN with only generated X-rays}} &\textbf{95.58\%} (0.90645,0.98364) &\textbf{93.38\%} (0.87809,0.96930) \\
\hline
\textit{\textit{Transfer GAN with Real and Generated covid X-rays (1400 = 159 real + 1241 generated) }} &\textbf{99.26\%} (0.95971,0.99981) &\textbf{99.26\%} (0.95971,0.99981)  \\
\hline
\textit{\textit{MTT-GAN with only generated X-rays}} &\textbf{96.32\%} (0.91629,0.98796) &\textbf{94.11\%} (0.88738,0.97426) \\
\hline
\textit{\textit{MTT-GAN with Real and Generated covid X-rays (1400 = 159 real + 1241 generated)}} &\textbf{99.26\%} (0.95971,0.99981) &\textbf{100\%} (1,1)  \\
\hline

\end{tabular}
\label{tab:binary}
\end{center}
\end{table}

\begin{figure*}[ht]
\centerline{\includegraphics[width=170mm,height=100mm,scale=10.5]{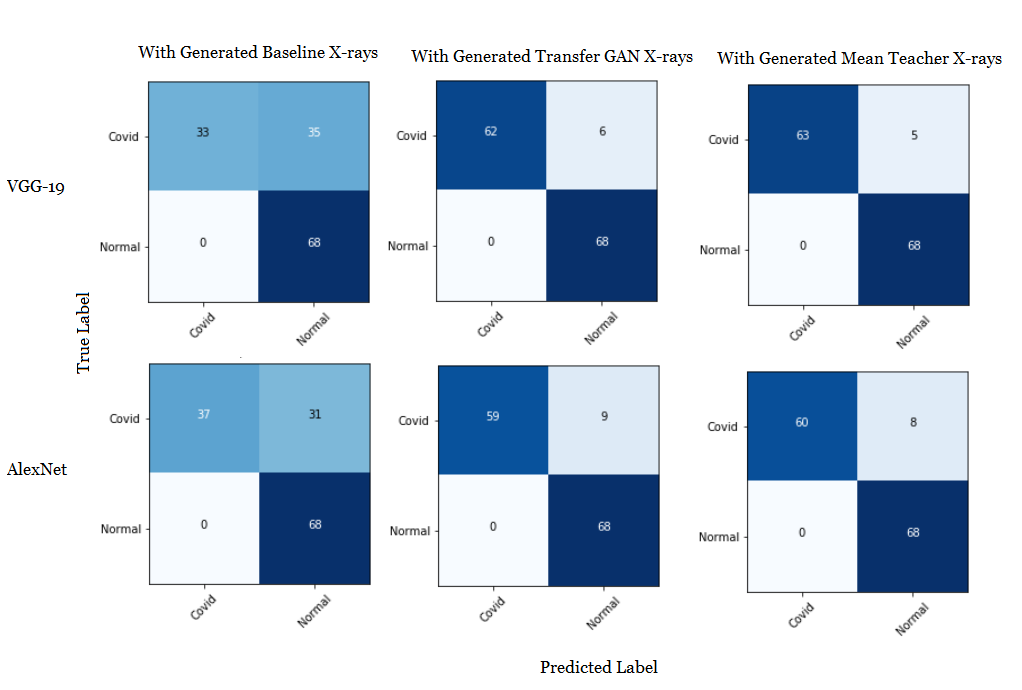}}
\caption{Binary Classification Confusion Matrix}
\label{fig:binary_matrix}
\end{figure*}

Quantitative analysis over the binary and multi-class classifiers demonstrate that MTT-GAN achieves the highest classification accuracy, and that both Transfer GAN and MTT-GAN achieve superior accuracy over the baseline GAN.  Furthermore the highest accuracy is achieved when a small amount of real imagery is included along with the generated COVID19 imagery.  Table \ref{tab:binary} shows the quantitative test accuracy for binary COVID19 vs normal classification. The first experiment, where the models were trained on the combination of baseline GAN images and the normal X-rays gave an accuracy of 74.26\% with a confidence interval of (0.66068,0.81374) for VGG-19  and 77.20\% (0.69233,0.83957) for AlexNet. In the second experiment, incorporating 159 real X-ray images increases this accuracy to 91.91\% (0.85989,0.95893) and 82.35\% (0.74890,0.88354) respectively.
For the Transfer GAN experiments the VGG-19 model gives an accuracy of 95.58\% (0.90645,0.98364) and the AlexNet gives us an accuracy of 93.38\% (0.87809,0.96930). with GAN images only.  The fourth experiment on binary classification consisted of 159 real covid X-rays, 1241 generated covid X-rays and 1400 real normal X-rays, and the VGG-19 model yielded a test accuracy of 99.26\% (0.95971,0.99981) and the AlexNet yielded an accuracy of 99.26\% (0.95971,0.99981).  Experiments five and six show that MTT-GAN further improves accuracy. Using only MTT-GAN augmentations gives us an accuracy of 96.32\% (0.91629,0.98796) and 94.11\% (0.88738,0.97426) respectively. Combining MTT-GAN agumentations with 159 real x-rays VGG-19 and AlexNet produce their highest accuracy of 99.26\% (0.95971,0.99981) and 100\% (1,1) respectively. Fisher tests show that the improvement in accuracy of MTT-GAN versus Baseline is significant with p-values $< 0.0001$ for all rows that compare MTT-GAN with Baseline Method without augmentation, and $p < 0.05$ for all rows that compare MTT-GAN with Baseline with augmentation except for VGG with only generated versus Baseline which is near significant with augmentation where $p = 0.064$.  Furthermore, the confusion matrices in figure \ref{fig:binary_matrix} demonstrate that MTT-GAN greatly increases the sensitivity relative to baseline GAN.

\begin{table}[htbp]
\label{MultiClass Classifier Experiments}
\caption{MultiClass Classifier Experiments}
\begin{center}
\begin{tabular}{|p{2.9cm}|p{2.1cm}|p{2.1cm}|}
\hline

\textit{\textit{Training Dataset  (5600 X-rays = 1400 covid X-rays+1400 normal X-rays+1400 bacterial pneumonia+1400 viral pneumonia)}} & \textit{\textit{VGG-19 Test Accuracy with confidence intervals  (272 X-rays = 68 covid X-rays+68 normal X-rays + 68 bacterial + 68 viral) }} & \textit{\textit{AlexNet Test Accuracy with confidence intervals (272 X-rays = 68 covid X-rays+68 normal X-rays + 68 bacterial + 68 viral) }} \\
\hline
\textit{\textit{Baseline Method (without augmentation) with only Generated Images}} &\textbf{76.10\%} (0.70582,0.81046) &\textbf{65.80\%} (0.59839,0.71430)  \\
\hline
\textit{\textit{Baseline Method (with augmentation) with only Generated Images}} &\textbf{79.41\%} (0.74113,0.84057) &\textbf{76.47\%} (0.70972,0.81382)  \\
\hline
\textit{\textit{Transfer-GAN method with only generated X-rays}} &\textbf{84.19\%} (0.79302,0.88317) &\textbf{82.72\%} (0.77693,0.87019) \\
\hline
\textit{\textit{Transfer-GAN method with Real and Generated covid X-rays (1400 = 159 real + 1241 generated)}} &\textbf{84.92\%} (0.80112,0.88961) &\textbf{83.89\%} (0.78899,0.87993)  \\
\hline
\textit{\textit{MTT-GAN with only generated X-rays}} &\textbf{83.45\%} (0.78496,0.87669) &\textbf{84.19\%} (0.79302,0.88317) \\
\hline
\textit{\textit{MTT-GAN with Real and Generated covid X-rays (1400 = 159 real + 1241 generated)}} &\textbf{84.93\%} (0.80112,0.88961) &\textbf{85.61\%} (0.80112,0.88961)  \\
\hline

\end{tabular}
\label{tab:multiclass}
\end{center}
\end{table}

\begin{figure*}
\centerline{\includegraphics[width=150mm]{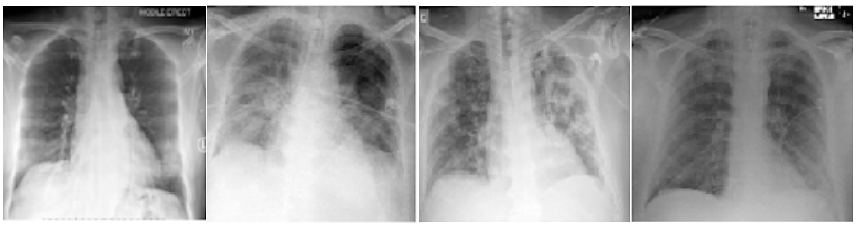}}
\caption{Real COVID19 X-rays}
\label{fig:real covid X-rays}
\end{figure*}

\begin{figure*}
\centerline{\includegraphics[width=150mm]{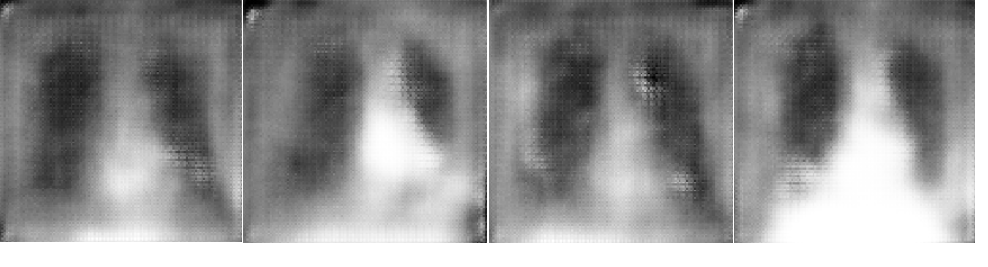}}
\caption{Baseline Model Generated X-rays}
\label{fig:baseline_generated}
\end{figure*}

\begin{figure*}
\centerline{\includegraphics[width=150mm]{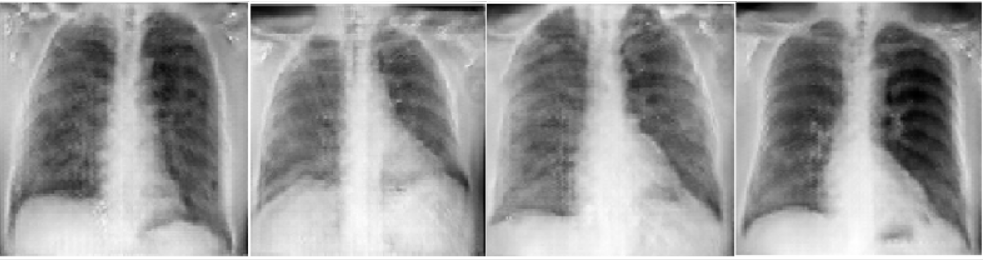}}
\caption{X-rays Generated using MTT-GAN}
\label{fig:improved_generated}
\end{figure*}

Table \ref{tab:multiclass} shows similar results for the multi-class classifiers. The baseline GAN with only generated images achieves the lowest results, where the VGG-19 model yields an accuracy of 76.10\% with a confidence interval of (0.70582,0.81046) and the AlexNet yields an accuracy of 65.80\% (0.59839,0.71430). In the second experiment for multi-class classification, using the baseline GAN with augmentation, the VGG-19 model yields an accuracy of 79.41\% (0.74113,0.84057) and the AlexNet yields an accuracy of 76.47\% (0.70972,0.81382).   When using images generated with the Transfer GAN, the accuracy is 84.19\% (0.79302,0.88317) using VGG-19 whereas the AlexNet yields an accuracy of 82.72\% (0.77693,0.87019). Combining the images generated by the Transfer GAN with the real images gives an accuracy of 84.92\% (0.80112,0.88961) using VGG-19 and 83.89\% (0.78899,0.87993) using AlexNet.  MTT-GAN with only generated X-rays achieved slightly lower accuracy for VGG19 of 83.45\% (0.78496,0.87669), but improved accuracy of 84.19\% (0.79302,0.88317) for AlexNet.  The highest accuracy was achieved by combining MTT-GAN with 159 real COVID19 X-rays yielding 84.93\% (0.80112,0.88961) accuracy for VGG19 and 85.61\% (0.80112,0.88961) accuracy for AlexNet.

\begin{figure*}[ht]
\centerline{\includegraphics[width=170mm,height=100mm,scale=10.5]{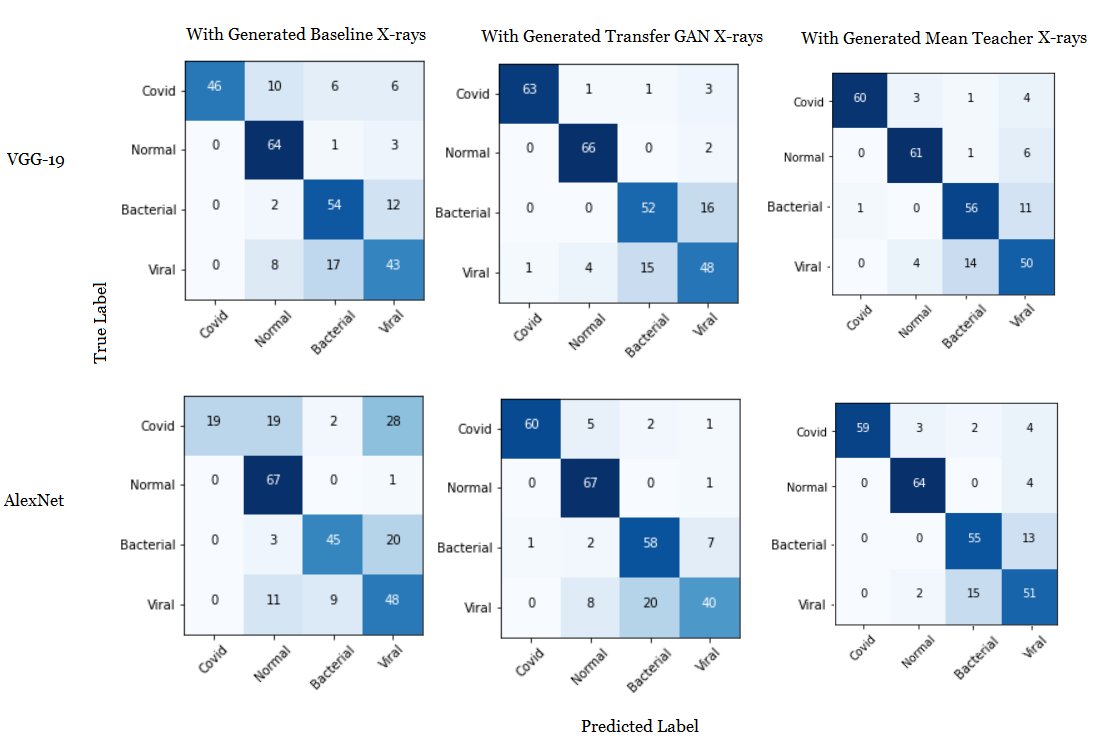}}
\caption{MultiClass Classification Confusion Matrix}
\label{fig:multiclass_matrix}
\end{figure*}

The confusion matrix in Fig. \ref{fig:multiclass_matrix} shows that the VGG-19 classifier is able to predict 46 out of 68 covid images accurately with the baseline generated X-rays whereas it predicts 63 of the 68 X-rays accurately using the X-rays generated by our MTT-GAN algorithm and architecture which is detecting 92.6\% of the covid cases accurately. Using the AlexNet, the baseline predicts only 19 out of the 68 positive covid cases, whereas with the MTT-GAN, 60 out of the 68 positive images are correctly identified.
Fisher tests show that the MTT-GAN improves accuracy relative to the baseline.  We achieve $p < 0.001$ for VGG comparisons of MTT-GAN versus Baseline without augmentation, and $p < 0.0001$ for equivalent Alex-net comparison.  Furthermore $p < 0.05$ for VGG comparison of MTT-GAN versus Baseline with augmentation, and $p < 0.01$ for equivalent Alex-net comparison.

\section{Qualitative Results}

Our qualitative results show that MTT-GAN greatly outperforms the baseline GAN, and generates images that approximate many anatomical features of the real images.  The real images are shown in Fig. \ref{fig:real covid X-rays}, while the baseline generated X-rays and the X-rays generated with transfer learning and mean teacher look are shown in Fig. \ref{fig:baseline_generated} and Fig. \ref{fig:improved_generated} respectively.

\begin{figure}[ht] 
\centerline{\includegraphics[width=60mm,scale=1.5]{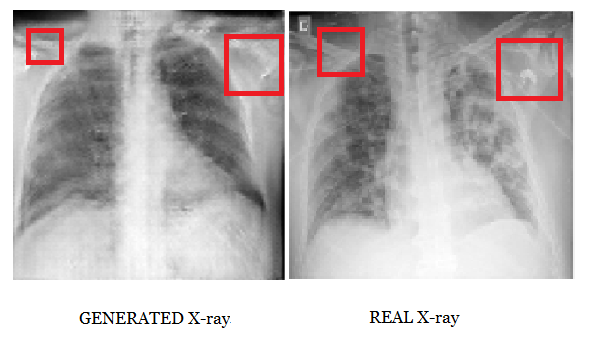}}
\caption{Survey Results by Radiologists}
\label{fig:survey}
\end{figure}

Although the MTT-GAN generated X-rays have many well defined anatomical features, they are nonetheless distinguishable from real X-rays by board-certified radiologists.  We conducted a survey with 2 radiologists where we displayed 25 pairs of real covid X-rays versus generated covid X-rays. We asked the radiologists to classify between the real and the generated X-rays and asked them to provide comments on the features that the generated X-rays have in comparison to the real X-rays. Both radiologists were able to correctly identify which image was real and which was fake in all 25 pairs.

The radiologists explained that the X-rays have greatly improved quality relative to the baseline, but fall short of diagnostic quality due to the following limitations: 1. low resolution (128x128) 2. systematic errors in the scapula and the clavicle bones as highlighted in Fig \ref{fig:survey}.    

The Radiologists suggested that a potential area of future work would be to incorporate skeletal background removal and/or style transfer methods to ensure that background features such as scapula and clavicle bone structures are consistent between generated and real images.

\section{Conclusion}

We present a novel MTT-GAN architecture for generating high quality synthetic chest X-ray images for patients with COVID19, and demonstrate improved the accuracy of binary and multi-class classifiers for automated COVID19 X-ray screening.  MTT-GAN addresses a notable challenge in that public datasets for COVID19 X-rays have highly limited data volumes.  To the best of our knowledge MTT-GAN is the first architecture to employ transfer learning from Kaggle Pneumonia to COVID19 for both the generator and discriminator models thereby greatly improving image quality.  This improved image quality translates to highly competitve COVID19 classification accuracy.  To the untrained eye MTT-GAN images appear similar to real COVID19 X-rays, although board certified radiologists can distinguish these images and suggest that more research is necessary to achieve diagnostic quality for human performance tasks.  Nevertheless, quality improvements to deep fakes are invaluable to improve classification accuracy for computer aided diagnosis. In conclusion, MTT-GAN is a novel approach that provides a notable improvement in the realism of generated deep fake COVID19 X-rays images.

\end{document}